\documentclass[11pt,a4paper]{article}
\usepackage[hyperref]{acl2020}
\usepackage{times}
\usepackage{latexsym}
\usepackage{graphicx}
\usepackage{algorithm}
\usepackage{amsmath}
\usepackage{multirow}
\usepackage{microtype}

\aclfinalcopy
\title{Extensively Matching for Few-shot Learning Event Detection}

\author{
Viet Dac Lai\textsuperscript{\rm 1}, Franck Dernoncourt\textsuperscript{\rm 2} and Thien Huu Nguyen\textsuperscript{\rm 1}\\
\textsuperscript{\rm 1}Department of Computer and Information Science, \\ University of Oregon, Eugene, Oregon, USA\\
\textsuperscript{\rm 2}Adobe Research, San Jose, CA, USA\\
{\tt \{vietl, thien\}@cs.uoregon.edu} \\
{\tt franck.dernoncourt@adobe.com}
}

\begin{document}
\maketitle
\begin{abstract}

Current event detection models under supervised learning settings fail to transfer to new event types. Few-shot learning has not been explored in event detection even though it allows a model to perform well with high generalization on new event types. In this work, we formulate event detection as a few-shot learning problem to enable to extend event detection to new event types. We propose two novel loss factors that matching examples in the support set to provide more training signals to the model. Moreover, these training signals can be applied in many metric-based few-shot learning models. Our extensive experiments on the ACE-2005 dataset (under a few-shot learning setting) show that the proposed method can improve the performance of few-shot learning.
\end{abstract}

\section{Introduction}

Event Detection (ED) is an important task in Information Extraction (IE) in Natural Language Processing (NLP). Event Detection is the task to detect event triggers from a given text (e.g. a sentence) and classify it into one of the event types of interest. The following sentence is an example of ED:

\textit{In 1997, the company} \textbf{hired} \textit{John D. Idol to take over as chief executive.}

In this example, an ideal event detection system should detect the word \textit{hired} as an event, and classify it to class of \textit{Personnel:Start-Position}, assuming that \textit{Personnel:Start-Position} is in the set of interested classes. 

The current works in ED typically employ traditional supervised learning based on feature engineering \cite{Li:14,Chen:17} and neural networks \cite{Nguyen:16a:joint,Chen:18,Lu:18}. The main problem with supervised learning models is that they can not perform well on unseen classes (e.g. training a model to classify daily events, then run this model to classify laboratory operations). As a result, supervised learning ED can not extend to unseen event types. A trivial solution is to annotate more data for unseen event types, then retraining the model with newly annotated data. However, this method is usually impractical because of the extremely high cost of annotation \cite{liu2019event}.

A human can learn about a new concept with limited supervision e.g. one can detect and classify events with 3-5 examples \cite{Grishman:05}. This motivates the setting we aim for event detection: \textbf{few-shot learning} (FSL). In FSL, a trained model rapidly learns a new concept from a few examples while keeping great generalization from observed examples \cite{vinyals2016matching}. Hence, if we need to extend event detection into a new domain, a few examples are needed to activate the system in the new domain without retraining the model. By formulating ED as FSL, we can significantly reduce the annotation cost and training cost while maintaining highly accurate results.

In a few shot learning iteration, the model is given a support set and a query instance. The support set consists of examples from a small set of classes. A model needs to predict the label of the query instance in accordance with the set of classes appeared in the support set. Typical methods employ a neural network to embed the samples into a low-dimension vector space \cite{vinyals2016matching,snell2017prototypical}, then, classification is done by matching those vectors based on vector distances \cite{vinyals2016matching,snell2017prototypical,sung2018relation}. One potential problem of prior FSL methods is that the model relies solely on training signals between query instance and the support set \cite{vinyals2016matching,snell2017prototypical,sung2018relation}. Thus, the matching information between samples in the support set has not been exploited yet. We believe that this is not an efficient use of training data because dataset in ED is very small \cite{Grishman:05}. Therefore, in this study, we propose to train an ED model using matching information (1) between query instance and the support set and (2) between the samples in the support themselves. This is implemented by adding two auxiliary factors into the loss function to constrain the learning process.

We apply the proposed training signals to different FSL models on the benchmark event detection dataset \cite{Grishman:05}. The experiments show that the training signal can improve the performance of the examined FSL models. To summarize, our contributions to this work include:

\begin{itemize}
    \item We formulate event detection as a few-shot learning problem to extend ED to new event types and provide a baseline for this new research direction. To our best knowledge, this is a new branch of research that has not been explored.
    \item We propose two novel training signals for FSL. These signals can remarkably improve the performance of existing FSL models. As these signals do not require any additional information (e.g. dependency tree or part-of-speech), they can be applied in any metric-based FSL models.
\end{itemize}

\section{Related work}

Early studies in event detection mainly address feature engineering for statistical models \cite{Ahn:06,Ji:08,Hong:11,Li:14,Li:15:improving} including semantic features and syntactic features. Recently, due to the advances with deep learning, many neural network architectures have been presented for ED, e.g. convolutional neural networks (CNN) \citep{Chen:15,Nguyen:15:event,Nguyen:16b:modeling,Nguyen:16c:twostage}, recurrent neural networks (RNN) \citep{Liu:17,Chen:18,Nguyen:16a:joint,nguyen:18:one} and graph convolutional neural networks (GCN) \cite{nguyen:18:graph,Pouran:19}. These methods formulate ED as a supervised learning problem which usually fails to predict the labels of new event types. 

By transitioning the symbolic event types to descriptive event types in the form of bags of keywords \cite{bronstein:15:seed,Peng:16:minimal,Lai:19}, the adaptibility of event detection can be formed as a supervised-learning problem. However, these studies have not examined FSL as we do in this work. One can also address this problem in zero-shot learning with data generated from abstract meaning representation \cite{huang:18:zeroshot} or two-stage pipeline ( trigger identification and few-shot event classification) based on dynamic memory network \cite{deng2020meta}. A recent study has employed few-shot learning for event classification \cite{lai2020exploiting}. Our work is similar in terms of formulation, however, we consider it in a larger extent of event detection where the \textit{NULL} event is also included.

Few-shot learning has been studied early in the literature \cite{thrun1996learning}. Before the era of the deep neural network, FSL approaches focused on building generative models that can transfer priors across classes. However, these methods are hard to apply to real applications because they require a subject-dedicated design such as handwritten characters \cite{lake2013one,wong2015one}. As a result, they cannot capture the nature of the distribution \cite{salimans2016improved}. Later studies, based on deep neural network, proposed metric learning to model the distribution of distance among classes, \cite{koch2015siamese} with many incremental improvements in distance functions such as cosine similarity \cite{vinyals2016matching}, Euclidean distance \cite{snell2017prototypical} and learnable distance function \cite{sung2018relation}. Metric-based FSL presents its advantages in two dimensions. First, it is based on the well-studied theory in distance functions. Second, the simplicity in architecture and training processes can encourage its application in practice. Recently, meta-learning with parameter update strategy is also proposed to enable the models to learn quickly in few training iterations \cite{santoro2016meta,finn2017model}. 

\section{Methodology}

Our goal in this work is to formulate ED as a FSL problem, which has not been done in prior work. In order to achieve this, this section is divided into three parts. In the section \ref{sec:formulation} we present the overall framework that formulate Event Detection as an Few-Shot Learning problem. Then, we present popular models for FSL in the prior work  and common sentence encoders which have been widely used in ED in section \ref{sec:framework}. Finally, we present two novel reguarlization technique to further improve the FSL model for ED in section \ref{sec:training-objective}.

\subsection{Event Detection as Few-shot Learning}
\label{sec:formulation}

In few-shot learning, models learn to predict the label of a query instance $x$ given a support set $S$ (a set of well-classified instances) and a set of classes $C$, which appears in the support set $S$. Prior studies in FSL employ $N$-way $K$-shot setting, in which there are $N$ clusters, which represent $N$ classes, each cluster contains $K$ data points (i.e., examples).

However, this setting is designed for problems that do not involve the ``NULL'' class (e.g., image classification and event classification). In event detection, the systems need to predict whether a query instance is an event (positive event type) or not (negative event type -- the ``NULL'' type) before it is further classified into one of the classes of interest. To this end, we propose to extend the N-way K-shot setting to be N+1-way K-shot setting. In this setting, the support set contains $N$ clusters representing $N$ positive event types and 1 cluster representing the NULL event type. The support set is denoted as follows:

\begin{equation*}
    \begin{aligned}
    S = & \{ (s^1_1, a^1_1, t_1),\ldots, (s^K_1, a^K_1, t_1), \\ 
        & \ldots \\
        & ( s^1_N, a^1_N, t_N),\ldots, ( s^K_N, a^K_N, t_N), \\
        & (s^1_{N+1}, a^1_{N+1}, t_{null}),\ldots, (s^K_1, a^K_{N+1}, t_{null})\}
    \end{aligned}
\end{equation*}
where:
\begin{itemize}
    \item $\{t_1, t_2, \cdots t_N\}$ is the set of positive labels, which indicate an event
    \item $t_{null}$ a special label for non-event.
    \item $(s^j_i, a^j_i, t_i)$ indicates that the $a^j_i$-th word in the sentence $s^j_i$ is the trigger word of an event mention with the event type $t_i$
\end{itemize}

\subsection{Framework}
\label{sec:framework}
Follow prior studies in FSL \cite{gao2019hybrid}, we employ the metric-based FSL framework with three components: instance encoder, prototype encoder, and classification module.

\subsubsection{Instance Encoder} 

Given a sentence of $L$ words $\{w_1,w_2,\cdots, w_L\}$ and the event mention $w_a$, which is the $a$-th word of the sentence, we first map discrete words to a continuous high dimensional vector space to facilitate neural network using both pre-trained word embedding and position embedding as follow:

\begin{itemize}
    \item In order to capture the syntactic and semantic of the word itself, we map each word in the sentence to a single vector using pre-trained word embedding, following previous studies in ED \cite{Nguyen:15:event}. After this step, we derive a sequence of vectors $\{e_1,e_2,\cdots, e_L\}$ where $e_i\in R^u$.
    
    \item To provide a sense of the relative position of a word regarding the position of the anchor word, we further provide position embedding. It is mapped from the relative distance, $i-a$, of the $i$-th word with respect to the anchor word, $a$-th word to a single vector $p_i\in R^v$. We randomly initialize this word embedding and update the embedding during the training process. 
    
    \item Following previous work \citep{Nguyen:15:event}, the final embedding of a word $w_i$ is derived by concatenating word embedding and position embedding $m_i=[e_i,p_i]\in R^{u+v}$.
\end{itemize}

Once we get the embedding for the whole sentence $E(s)=\{m_1, m_2, \cdots, m_L\}$, we employ a neural network, denoted as $f$, to encode the information of an instance $(s,a)$ of the anchor $w_a$ under the context in the sentence $s$ into a single vector $v=f(E(s),a)$. In this work, consider the three following neural network architectures for this encoding purpose:

\begin{itemize}
    \item Convolution Neural Network (CNN) \cite{Kim:14} encodes the sentence by convolution operation on $k$ consecutive vectors representing $k$-gram. Follow \cite{Nguyen:15:event}, we use multiple kernel sizes $k\in\{2,3,4,5\}$ to cover the context with 150 filters for each kernel size. To squeeze the information of the sentence, we apply max pooling to the top convolution layer to get a pooled vector $p$. We also introduce local embedding $e_{[a-w,a+w]}$ with window size $w=2$. We concatenate pooled vector and local embeddings, and feed them through multiple dense layer to get the final representation:
    
    $$
    v=W[p,e_{[a-w,a+w]}]
    $$
    
    \item Long Short-Term Memory (LSTM) \cite{Hochreiter:97}, at each step $i$, computes a hidden vector $h_i$ from the hidden vector of the previous step $h_{i-1}$ and the current input vector $e_i$. To capture the context from both sides a word in the sentence, we employ two separate LSTMs running on forward and backward directions. Eventually, we can obtain two sequence of hidden vector $\{h^{forward}_i, \cdots, h^{forward}_L\}$ and $\{h^{backward}_i, \cdots, h^{backward}_L\}$. Finally, we concatenate the $a$-th vectors, at the position of the anchor, to form the representation of the instance:
    $$v=\text{concat}(h^{forward}_a, h^{backward}_a)$$

    \item Graph Convolutional Neural Network features graph convolution \cite{Kipf:17} on syntactic dependency graph, which allows the model to access to the nonconsecutive words based on the connection on the syntactic dependency tree. Following  \cite{nguyen:18:graph}, we transform the dependency tree into a syntactic graph by making it an undirected graph and adding node loops. The hidden vectors $h^l_i$ of the $l$-th vector is obtained by feeding hidden vectors of the $l-1$-th layer through a GCN layer \cite{Kipf:17}.
    The final representation is the hidden vector in the top layer at the position of the trigger $h_a^L$ where $L=2$ is the number of GCN layers.
    
\end{itemize}

\subsubsection{Prototype Encoder} 

This module computes a representative vector, called \textbf{prototype}, for each class $t\in T$ in the support set $S$ from its instances' vectors. We employ two variants of prototype computation. 

The first version, proposed in the original Prototypical Network \cite{snell2017prototypical}, considers all representation vectors are equally important. To calculate the prototype for a class $t_i$, it aggregates all the representation vectors of the instance of class $t_i$, and then perform averaging over all vectors :

\begin{equation}
\textbf{c}_i = \frac{1}{K}\sum_{(s^j_i, a^j_i, t_i) \in S} f(E(s^j_i), a^j_i)
\label{eq:c1}
\end{equation}

On the other hand, it was claimed that the supporting vectors are conditionally important with respect to the query $(q,p)$. Thus, the second version computes the prototype as a weighted sum of the supporting vectors. The weights are obtained by attention mechanism according to the representational vector of the query as follow:

\begin{equation}
\begin{aligned}
    \textbf{c}_i &= \sum_{(s^j_i, a^j_i, t_i)\in S} \alpha_{ij} f(E(s^j_i), a^j_i) \\
\end{aligned}
\label{eq:c2}
\end{equation}
where
\begin{equation*}
\begin{aligned}
    \alpha_{ij}&=\frac{\text{exp}(b_{ij})}{\sum_{(s^k_i, a^k_i, t_i)\in S} \text{exp}(b_{ik})}\\
    b_{ij} &= \sum\Big[\sigma(f(E(s^j_i), a^j_i) \odot f(E(q),p))\Big]
    \\
    \odot &\text{ denotes the element-wise product.}
\end{aligned}
\end{equation*}

\begin{table*}
\centering
\begin{tabular}{|l|c|c|c|c|c|c|}
    \hline
    
    Model & \multicolumn{3}{|c|}{5+1-way 5-shot} & \multicolumn{3}{|c|}{10+1-way 10-shot}\\
    \hline
    Encoder & CNN & LSTM & GCNN & CNN & LSTM & GCNN \\
    \hline
    Proto& 70.85& 68.77& 71.30&  61.43& 57.89& 62.36\\
    Proto+Att& 71.23& 69.32& 72.76&  63.50& 59.56& 65.08\\
    Relation& 54.36& 68.33& 58.37&  41.37& 62.85& 44.43\\
    Matching& 34.71& 49.40& 32.49&  23.05& 33.84& 21.51\\
    \hline
\end{tabular}
\caption{F1-score (micro) of models using CNN, LSTM and GCN encoders without proposed losses.}
\label{tab:ed-cnn}
\end{table*}

\subsubsection{Classification Module}

This module computes the distribution on all the event types $T$ of a query instance $x=(q,p)$ using a distance/similarity function $d: R\leftarrow R^d$.

\begin{equation}
P(y=t_i|x,S) = \frac{\text{exp}(-d(v, \textbf{c}^i))}{\sum^N_{j=1}\text{exp}(-d(v, \textbf{c}^j))}
\end{equation}

where $d$ is a distance/similarity function, $v=f(q,p)$ is the representation vector of the given query instance and $\textbf{c}^i$ and $\textbf{c}^j$ are the prototype vectors obtained in either Equation (\ref{eq:c1}) or Equation (\ref{eq:c2}) from the support set $S$.

In this paper, we examine three kinds of distance/similarity function with prototype module to form 4 model as follow:

\begin{itemize}
    \item Cosine similarity with averaging prototype as \textbf{Matching} network \cite{vinyals2016matching}.
    \item Euclidean distance with averaging prototype as \textbf{Proto} network \cite{snell2017prototypical}.
    \item Euclidean distance with weighted sum prototype as \textbf{Proto+Att} network \cite{gao2019hybrid}.
    \item Learnable distance function with averaging prototype as \textbf{Relation} network \cite{sung2018relation}.
\end{itemize}
\subsection{Training Objectives}
\label{sec:training-objective}
In the literature, a metric-based FSL model is typically trained by minimizing the negative log-likelihood as follow:

\begin{equation}
    L_{query}(x, S) = - \log P(y = t|x,S)
    \label{eq:l1}
\end{equation}
where $x$, $t$, $S$ are query instance, ground truth label, and support set, respectively.

This loss function exploits the signal of matching information between the query instance and the supporting instances. It can work efficiently in computer vision because the number of samples in computer vision datasets are typically huge. However, in NLP tasks, the dataset is commonly relatively much smaller (e.g. ACE 2005 contains 4000 positive examples). So using this loss function is not enough to deliver a good system. 

Therefore, providing more training signals is crucial to the problem which involves a small dataset. Fortunately, the support set is a well-classified set of instances with $K$ examples per class in a total of $N$ classes. In this paper, we proposed two ways to exploit this resourceful set as follow:

\begin{itemize}
    \item Intra-cluster matching: We argue that the representational vectors in the same class should be close to each other. Therefore, we minimize the distance between instance in the same class.
    \begin{equation}
        L_{intra} = \sum^{N}_{i=1}\sum^{K}_{k=1}\sum^{K}_{j=k+1}\text{mse}(v^j_i, v^k_i)
    \label{eq:l2}
    \end{equation}
    \item Inter-cluster information: We also argue that the clusters should distribute far away from each other. Hence, their prototypes are also distant from the other. Hence, we maximize the distances between pairs of prototypes.
    \begin{equation}
        L_{inter} =1 - \sum^{N}_{i=1}\sum^{N}_{j=i+1}\text{cosine}(c_i, c_j)
        \label{eq:l3}
    \end{equation}
\end{itemize}

In this work, we train our model using a combination of the loss functions in equations \ref{eq:l1}, \ref{eq:l2},\ref{eq:l3}. We control the contribution of the additional losses by two hyperparameters $\beta$ and $\gamma$ as follow:

\begin{equation}
    L = L_{query} + \beta \hat{L}_{intra} + \gamma \hat{L}_{inter}
    \label{eq:l4}
\end{equation}
where $\hat{L}_{intra}$ and $\hat{L}_{inter}$ are scaled losses with respect to $L_{query}$, and $\beta$ and $\gamma$ are the trade-off parameters.


\section{Experiments}
\begin{table*}
\centering
\begin{tabular}{|l|l|c|c|c|c|}
    \hline
    \multirow{2}{*}{Encoder} & \multirow{2}{*}{Model} & \multicolumn{2}{|c|}{5+1-way 5-shot} & \multicolumn{2}{|c|}{10+1-way 10-shot} \\
    \cline{3-6}
    & & Original & + $L_{inter}+ L_{intra}$ & Original & $+L_{inter}+ L_{intra}$ \\
    \hline
    CNN & Proto & 70.85 & \textbf{72.07} & 61.43 & \textbf{62.84}\\
    LSTM & Proto & 68.77 & \textbf{78.09} & 57.89 & \textbf{72.78}\\
    GCN & Proto & 71.30 & \textbf{71.82} & 62.36 & \textbf{63.49}\\
    \hline
    CNN & Proto+Att & 71.23 & \textbf{72.46} & 63.5 & \textbf{64.38}\\
    LSTM & Proto+Att & 69.32 & \textbf{78.44} & 59.56 & \textbf{72.94}\\
    GCN & Proto+Att & 72.76 & \textbf{72.92} & 65.08 & \textbf{66.10}\\
    \hline
\end{tabular}
\caption{F1-score (micro) of models using CNN, LSTM, and GCN. \textit{Original} columns show the models without additional training signal.  $L_{inter}+ L_{intra}$ columns demonstrate the models with additional inter and intra loss functions.}
\label{tab:ed-distance}
\end{table*}

\subsection{Data}
We use the ACE-2005 dataset to evaluate all of the models in this study. ACE-2005 is a benchmark dataset in event detection with 33 positive event subtypes, which are grouped into 8 event types \textit{Business, Contact, Conflict, Justice, Life, Movement, Personnel, and Transaction}. Although the dataset is split into training, development, and testing sets, we cannot use these splits directly because, in FSL, the set of event types in the training set and testing sets are disjoint. Therefore, we further split these datasets to satisfy three conditions for FSL:
\begin{itemize}
    \item The set of event types in the training set $T^{train}$ are disjoint to those in the development and the testing set:
    
    $T^{dev} \equiv T^{test}; T^{train} \cap T^{test}=\emptyset $;
    
    \item In order to run FSL with the 10-way 10-shot setting, the set of event subtypes should contain at least 10 subtypes.
    \item The training set should contain as many samples as possible.
\end{itemize}

Based on these criteria, we use all samples belonging to 4 event types: \textit{Business, Contact, Conflict} and \textit{Justice} as the training set. While the rest (\textit{Life, Movement, Personnel} and \textit{Transaction}) are used for the development and testing sets. We split the sample by ratio 50:50 in every subtype to ensure the balance of the development and the testing set. Finally, since there are event types that have less than 15 examples, we eliminate all of these from the training, development, and testing set.

\subsection{Hyper-parameters}

We evaluate using 5+1-way 5-shot and 10+1-way 10-shot FSL settings. Although it was seen that the higher number of classes we have during the training time, the better performance on testing \cite{snell2017prototypical}, we avoid feeding all event types in every iteration during training time. We manage to sample 20 positive classes (over 21 in the training set) in each training iteration.


We initialize the embedding vectors with 300-dimension GLoVe embedding, trained from 6 billion tokens. We use 50-dimension position embedding and initialize it randomly. These embedding vectors are updated during training time.

We train Proto, Proto+Att, and Matching using Stochastic Gradient Decent (SGD) optimizer while Relation is trained with AdaDelta optimizer because SGD hardly converges with Relation network. The learning rate is initialized to 0.03 and decays after every 500 iterations. We trained our models in 2500 iterations and evaluation at every 200 iterations. 

In order to find the best set of $\beta$ and $\gamma$, we do grid search with with $(\beta,\gamma)\in\{0.0, 0.1, 0.2,0.3\}^2$.

\subsection{Result}

\begin{table*}
\centering
\begin{tabular}{|l|l|c|c|c|c|}
    \hline
    Encoder & FSL Model & Original & +Inter & +Intra & +Intra+Inter \\
    \hline
    CNN & Proto & 67.92 & 68.78 & 68.83 & 69.37 \\
    LSTM & Proto & 65.94 & 65.28 & 72.07 & 77.56 \\
    GCN & Proto & 69.28 & 70.05 & 69.49 & 70.11 \\
    \hline
    CNN & Proto+Att & 69.90 & 70.23 & 70.06 & 70.43\\
    LSTM & Proto+Att & 67.26 & 67.48 & 72.00 & 77.81 \\
    GCN & Proto+Att & 71.65 & 71.75 & 71.56 & 71.18\\
    \hline
\end{tabular}
\caption{Ablation study: F1-score (micro) of Prototypical-based models on dev set with 5+1-way 5-shot FSL setting}
\label{tab:ed-ablation}
\end{table*}

In this section, we perform our experiment in three steps:(1) find the best FSL models among Proto, Proto+Att, Matching, Relation models; (2) evaluate the proposed additional training factors and (3) analyze the effectiveness of each training factor in an ablation study.

Table \ref{tab:ed-cnn} shows the F-scores of four models using three kinds of sentence encoders on the ACE-2005 dataset under 5+1-way 5-shot and 10+1-way 10-shot FSL settings without our proposed losses. As can be seen from the Table \ref{tab:ed-cnn}, the performance of the models on 5+1-way 5-shot is always better than 10+1-way 10-shot because the number of classes needs to be classified in the 10+1-way setting is almost twice as much of in 5+1-way setting. Second, we can see that Prototypical-based (Proto and Proto+Att) models outperform the Matching network and the Relation network on both FSL settings. Among Prototypical network models, Proto+Att is slightly better than Proto with a 0.8\% performance gap in the 10+1-way 10-shot setting.

Most importantly, Table \ref{tab:ed-distance} presents the F-scores of Proto and Proto+A with the proposed loss functions (i.e., $L_{intra}$, $L_{inter}$). As we can see from the table, the proposed loss functions can significantly improve the performance of Proto and Proto+Att models over different encoders (i.e., CNN, LSTM, and GCN), clearly demonstrating the benefits of the intra and inter-similarity constraints in this work.

\subsection{Ablation Study}

In this study, we introduce two penalization factors, presented in Equations \ref{eq:l2} and \ref{eq:l3}. 

Besides the FSL formulation for event detection, a major contribution in this work involves the two loss functions $L_{intra}$ and $L_{inter}$ to improve the representation vectors for the models. To evaluate the contribution of these terms, Table \ref{tab:ed-ablation} shows the performance of the FSL models with different combinations of loss functions on the development set. In particular, we focus on the prototypical-based FSL model on the 5+1-way 5-shot setting in this analysis (although the similar trends of the performance are also observed for the other models and settings). The ``\textit{Original}'' column corresponds to the models where both $L_{inter}$ and $L_{intra}$ are not applied. The other columns, on the other hand, report the performance of the models when the combinations $L_{inter}$, $L_{intra}$, and $L_{inter}+ L_{intra}$ of the loss terms are introduced.



It is clear from the table that both loss terms are important for the FSL models for ED as eliminating any of them would significantly hurt the performance excepting the Proto+Att model with GCN encoder. The best performance is achieved with both loss terms are applied, thus testifying to the benefits of the proposed regularization techniques in this work.


\section{Conclusion}

In this paper, we address the problem of extending event detection to unseen event types through few-shot learning. We investigate four metric-based few-shot learning models with different encoder types (CNN, LSTM, and GCN). Moreover, we propose two novel loss functions to provide more training signals to the model exploiting domain-matching information in the support set. Our extensive experiments show that our method increases the efficiency of using training data, resulting in better classification performance. Our ablation study shows that both intra-cluster matching and inter-cluster matching contributes to the improvement.

\section*{Acknowledgments}

This research is based upon work supported in part by the Office of the Director of National Intelligence (ODNI), Intelligence Advanced Research Projects Activity (IARPA), via IARPA Contract No. 2019-19051600006 under the Better Extraction from Text Towards Enhanced Retrieval (BETTER) Program. The views and conclusions contained herein are those of the authors and should not be interpreted as necessarily representing the official policies, either expressed or implied, of ODNI, IARPA, the Department of Defense, or the U.S. Government. The U.S. Government is authorized to reproduce and distribute reprints for governmental purposes notwithstanding any copyright annotation therein. This document does not contain technology or technical data controlled under either the U.S. International Traffic in Arms Regulations or the U.S. Export Administration Regulations.

\bibliographystyle{acl_natbib}
\bibliography{ref}

\end{document}